\documentclass[runningheads, envcountsame, a4paper]{llncs}
\usepackage{graphicx}
\usepackage{bmpsize}
\usepackage{amsmath}
\usepackage[linesnumbered,ruled,vlined]{algorithm2e}
 \usepackage{microtype}
\usepackage{lmodern}

\title{Efficient Geometric-based Computation of the String Subsequence Kernel\thanks{This work  is supported by the MESRS - Algeria under Project 8/U03/7015.}}
\titlerunning{Efficient Geometric-based Computation of the String Subsequence Kernel}
\author{Slimane Bellaouar\inst{1} \and Hadda Cherroun\inst{1} \and Djelloul Ziadi\inst{2}}
\authorrunning{S. Bellaouar, H. Cherroun, and D. Ziadi}
\tocauthor{Slimane~Bellaouar, Hadda~Cherroun, and Djelloul~Ziadi}
\institute{Laboratoire LIM, Universit\'{e} Amar Telidji, Laghouat, Alg\'{e}rie
\email{$\{$s.bellaouar,hadda\_cherroun$\}$@mail.lagh-univ.dz}
\and Laboratoire LITIS - EA 4108, Universit\'{e} de Rouen, Rouen,  France\\
\email{djelloul.ziadi@univ-rouen.fr}}
\date{}

\begin{document}
\maketitle
\setcounter{footnote}{0}
\begin{abstract}
Kernel methods are powerful tools in machine learning. They have to be computationally efficient. In this paper, we present a novel Geometric-based approach to compute efficiently the string subsequence kernel (SSK). Our main idea is that the SSK computation reduces to range query problem. We started by the construction of a \textit{match list} $L(s,t)=\{(i,j):s_{i}=t_{j}\}$ where $s$ and $t$ are the strings to be compared; such match list contains only the required data that contribute to the result. To compute efficiently the SSK, we extended the  \textit{layered range tree} data structure to a \textit{layered range sum tree}, a range-aggregation data structure. The whole process takes $ O(p|L|\log|L|)$ time and $O(|L|\log|L|)$ space, where $|L|$ is the size of the match list and $p$ is the length of the SSK. We present empiric evaluations of our approach against the dynamic and the sparse programming approaches both on synthetically generated data and on newswire article data. Such experiments show the efficiency of our approach for large alphabet size except for very short strings. Moreover, compared to the sparse dynamic approach, the proposed approach outperforms absolutely for long strings.
\end{abstract}

\section{Introduction}
Kernel methods \cite {Cristianini:1999:ISV:345662} offer an alternative solution to the limitation of traditional machine learning algorithms, applied solely on linear separable problems. They map data into a high dimensional feature space where we can apply linear learning machines based on algebra, geometry and statistics. Hence, we may discover non-linear relations. Moreover, kernel methods enable other data type processings (biosequences, images, graphs, \dots).

Strings are among the important data types. Therefore, machine learning community devotes a great effort of research to string kernels, which are widely used in the fields of bioinformatics and natural language processing. The philosophy of all string kernels can be reduced to different ways to count common substrings or subsequences that occur in both strings to be compared, say $s$ and $t$.

In the literature, there are two main approaches to improve the computation of the SSK. The first one is based on dynamic programming; Lodhi et al. \cite {Lodhi:2002:TCU:944790.944799} apply dynamic programming paradigm to the suffix version of the SSK. They achieve a complexity of $O(p|s||t|)$, where $p$ is the length of the SSK. Later, Rousu and Shawe-Taylor \cite{Rousu:2005:ECG:1046920.1088717} propose an improvement to the dynamic programming approach based on the observation that most entries of the dynamic programming matrix (DP) do not really contribute to the result. They use a set of match lists combined with a sum range tree. They achieve a complexity of  $O(p|L|\log{\min(|s|, |t|)})$, where $L$ is the set of matches of characters in the two strings. Beyond the dynamic programming paradigm, the trie-based approach \cite{LeslieEtAl:03,Rousu:2005:ECG:1046920.1088717,Shawe-Taylor:2004:KMP:975545}  is based on depth first traversal on an implicit trie data structure. The idea is that each node in the trie corresponds to a co-occurrence between strings. But the number of gaps is restricted, so the computation is approximate.

Motivated by the efficiency of the computation, a key property of kernel methods, in this paper we focus on improving the SSK computation. Our main idea consists to map a machine learning problem on a computational geometry one. Precisely, our geometric-based SSK computation reduces to 2-dimensional range queries on a layered range sum tree (a layered range tree that we have extended to a range-aggregate data structure). We started by the construction of a \textit{match list} $L(s,t)=\{(i,j):s_{i}=t_{j}\}$ where $s$ and $t$ are the strings to be compared; such match list contains only the required data that contribute to the result. To compute efficiently the SSK, we constructed a \textit{layered range sum tree} and applied the corresponding computational geometry algorithms. The overall time complexity is $O(p|L|\log{|L|})$, where $|L|$ is the size of the match list.

The rest of this paper is organized as follows. Section~\ref{Preliminaries} deals with some concept definitions and introduces the layered range tree data structure. In section~\ref{SSK}, we recall formally the SSK computation. We also review three efficient computations of the SSK, namely, dynamic programming, trie-based and sparse dynamic programming approaches. Our contribution is addressed in Section~\ref{LRST}.  Section~\ref{Experimentation} presents the conducted experiments and discusses the associated results, demonstrating the practicality of our approach for large alphabet sizes. Section~\ref{Conclusions} presents conclusions and further work.

\section{Preliminaries}
\label {Preliminaries}
We first deal with concepts of string, substring, subsequence and kernel. We then present the layered range tree data structure.
\subsection {String} 
Let $ \Sigma  $ be an alphabet of a finite set of symbols. We denote the number of symbols in $ \Sigma $ by $ |\Sigma| $. A string $s = s_{1}...s_{|s|}$ is a finite sequence of symbols of length $|s|$ where $s_{i}$ marks the $i^{th}$ element of $s$. The symbol $\epsilon$ denotes the empty string. We use $\Sigma^{n}$ to denote the set of all finite strings of length $n$ and $\Sigma^{*}$ the set of all strings.
The notation $[s=t]$ is a boolean function that returns
\[
\left\{
\begin {array}{r c l}
1 & &  \textrm {if $s$ and $t$ are identical;} \\
0 & &  \textrm {otherwise.}
\end {array}
\right.
\]

\subsection {Substring}
For $1\le i \le j \le |s|$, the string $s(i:j)$ denotes the substring $s_{i} s_{i+1}...s_{j}$ of $s$. Accordingly, a string $t$ is a substring of a string $s$ if there are strings $u$ and $v$ such that $s=utv$ ($u$ and $v$ can be empty). The substrings of length $n$ are referred to as $n$-grams (or $n$-mers).

\subsection {Subsequence}
The string $t$ is a subsequence of $s$ if there exists an increasing sequence of indices $I=(i_{1},...,i_{|t|})$ in $s$, $(1\le i_{1}<...<i_{|t|}\le |s|) $ such that $t_{j}=s_{i_{j}}$, for $j=1,...,|t|$.
In the literature, we use $t=s(I)$ if $t$ is a subsequence of $s$ in the positions given by $I$. The empty string $\epsilon$ is indexed by the empty tuple. The absolute value $|t|$ denotes the length of the subsequence $t$ which is the number of indices $|I|$, while $l(I)=i_{|t|}-i_{1}+1$ refers to the number of characters of $s$ covered by the subsequence $t$.

\subsection {Kernel methods}
Traditional machine learning and statistic algorithms have been focused on linearly separable problems (i.e. detecting linear relations between data). It is the case where data can be represented by a single row of table. However, real world data  analysis requires non linear methods. In this case, the target concept cannot be expressed as simple linear combinations of the given attributes \cite {Cristianini:1999:ISV:345662}. This was highlighted in 1960 by Minsky and Papert.

Kernel methods were proposed as a solution by embedding  the data in a high dimensional feature space where linear learning machines based on algebra, geometry and statistics can be applied.
This embedding is called kernel. It arises as a similarity measure (inner product) in a high dimension space so-called feature description.

The trick is to be able to compute this inner product directly from the original data space using the kernel function. This can be formally clarified as follows: the kernel function $K$ corresponds to the inner product in a feature space $F$ via a map $\phi$.

\begin{eqnarray*}
\phi&  :& X \to F \\
&&x \mapsto \phi (x)\\
K(x,x')&=& \langle \phi (x) , \phi(x')\rangle.
\end{eqnarray*}

\subsection {Layered Range Tree}
\label{lrt}
\begin{figure}[h]
\centering
\includegraphics[height=65mm,width=120mm]{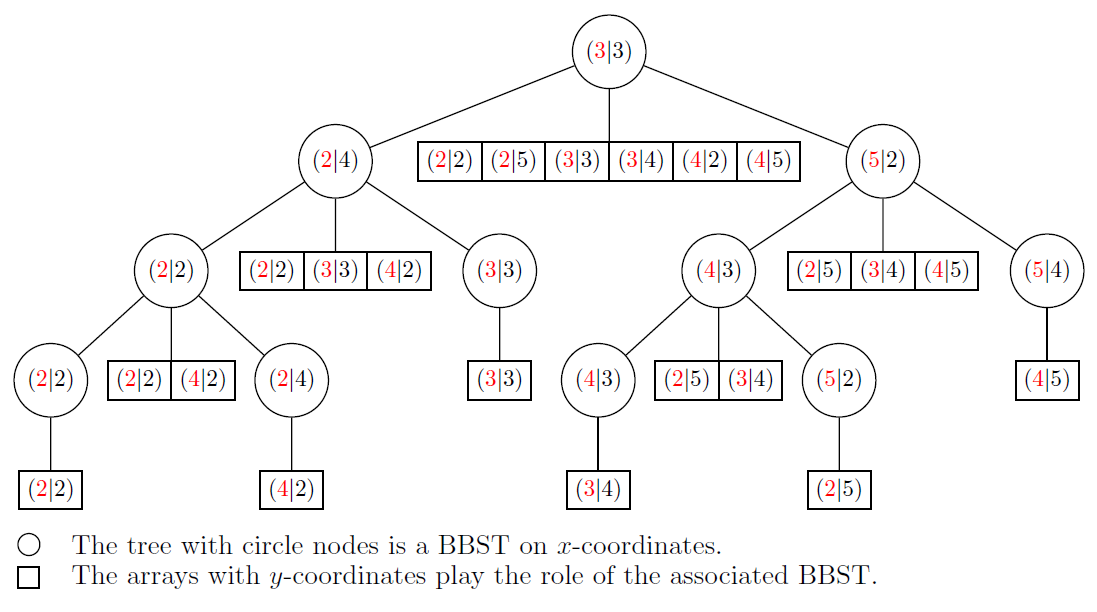}
\caption{ A 2-dimensional range tree.}
\label{RT}
\end{figure}

A Layered Range Tree (LRT) is a spatial data structure that supports orthogonal range queries. It is judicious to describe a 2-dimensional range tree inorder to understand LRT.
Consider a set $S$ of points in $\mathcal{R}^2$. A range tree is primarily a balanced binary search tree (BBST) built on the  $x$-coordinate of the points of $S$. Data are stored in the leaves only.  Every node $v$ in the BBST is augmented by an associated data structure ($\mathcal{T}_{assoc}(v)$) whitch is a 1-dimensional range tree, it can be a BBST or a sorted array, of a canonical subset $P(v)$ on $y$-coordinates. The subset $P(v)$ is the  points stored in the leaves of the sub tree rooted at the node $v$. Figure~\ref{RT} depicts a 2-dimensional range tree for a set of points $S = \{(2,2), (5,2), (3,3), (4,3), (2,4), (5,4)\}$. In the case where two points have the same $x$ or $y$-coordinate, we have to define a total order by using a lexicographic one. It consists to replace the real number by a composite-number space~\cite{Berg:2008:CGA:1370949}. The composite number of two reals $x$ and $y$ is denoted by $(x|y)$, so for two points, we have:
 \[(x|y) < (x{'}|y{'}) \Leftrightarrow x < x{'}  \lor (x=x{'} \land y<y{'}).\]
Based on the analysis of computational geometry algorithms, our 2-dimensional range tree for a set $S$ of $n$ points requires $O(n\log n)$ storage and can be constructed in $O(n\log n)$ time. 

The range search problem consists to find all the points of $S$ that satisfy a range query. A useful idea, in terms of efficiency, consists on treating a rectangular range query as a two nested 1-dimensional queries. In other words, let $[x_{1}:x_{2}]\times[y_{1}:y_{2}]$ be a 2-dimensional range query, we first ask for the points with $x$-coordinates in the given 1-dimensional range query $[x_{1}:x_{2}]$. Consequently,  we select a collection of $O(\log n)$ subtrees. We consider only the canonical subset of the resulted subtrees, which contains, exactly, the points that lies in the $x$-range $[x_{1}:x_{2}]$.
At the next step, we will only consider the points that fall in the $y$-range $[y_{1}:y_{2}]$.\\
The total task of a range query can be performed in $O(\log^{2}n+k)$ time, where $k$ is the number of points that are in the range. We can improve it by enhancing the 2-dimensional range tree with the fractional cascading technique which is described in the following paragraph. 

The key observation made during the invocation of a rectangular range query is that we have to search the same range	$[y_{1}:y_{2}]$ in the associated structures  of $O(\log n)$ nodes found while querying the main BBST by the range query $[x_{1}:x_{2}]$. Moreover, there exists an inclusion relationship between these associated structures. The goal of the fractional cascading consists on executing the binary search only once and use the result to speed up other searches without expanding the storage by more than a constant factor.

\begin{figure}[h]
\centering
\includegraphics[height=65mm,width=120mm]{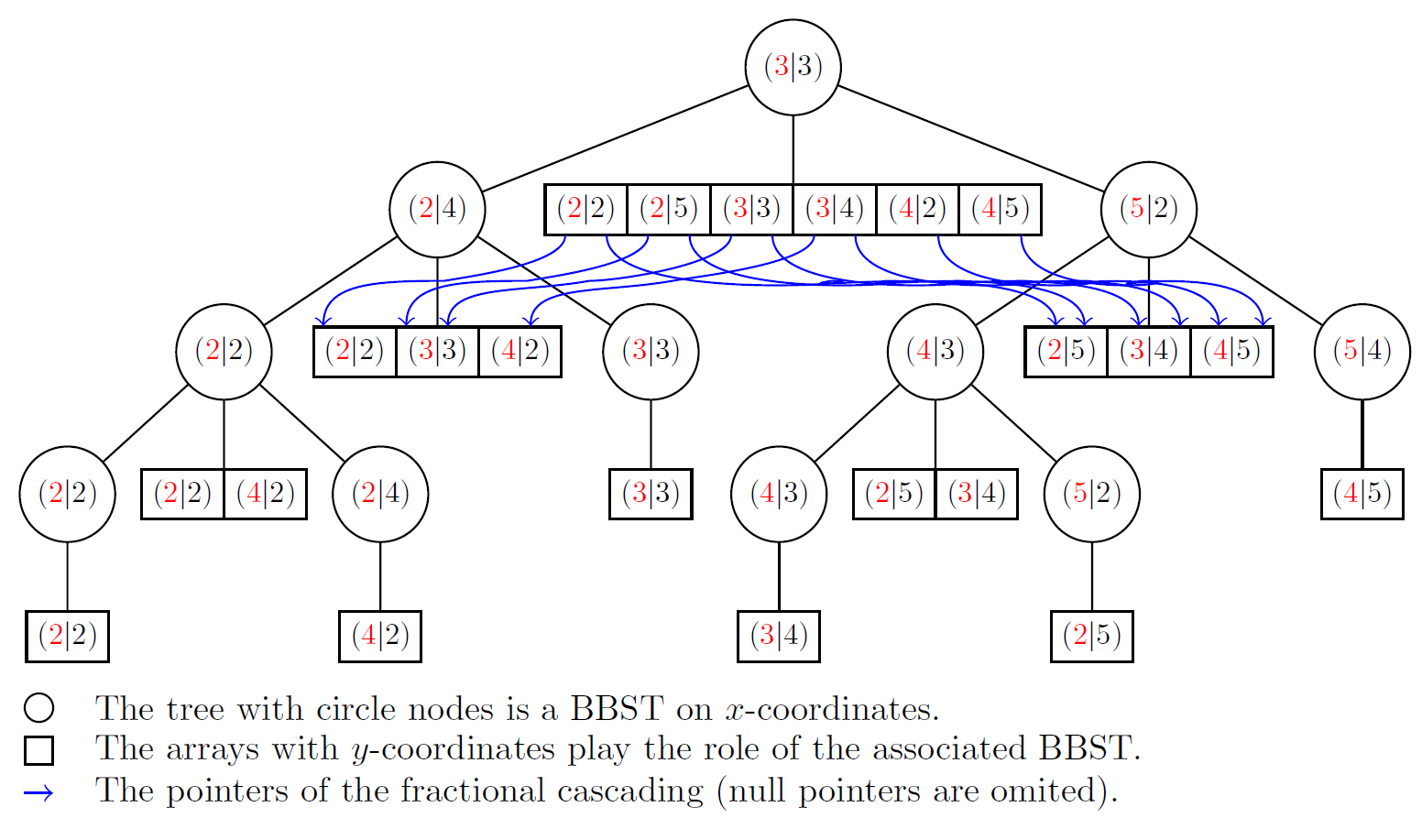}
\caption{A layered range tree, an illustration of the fractional cascading (only between two levels).}
\label{LRT}
\end{figure}

The application of the fractional cascading technique introduced by \cite {DBLP:/algorithmica/ChazelleG86} on a range tree creates a new data structure so called \textit{layered range tree}. The technique consists to add pointers from the entries of an associated data structure $\mathcal{T}_{assoc}$ of some level  to the entries of an associated data structure below, say $\mathcal{T'}_{assoc}$  as follows: If $\mathcal{T}_{assoc}[i]$ stores a value with the key $y_{i}$, then we store a pointer to the entry in $\mathcal{T'}_{assoc}$ with the smallest key larger than or equal $y_{i}$.  We illustrate such technique  in Fig.~\ref{LRT} for the same set represented in  Fig.~\ref{RT}.
Using this technique, the rectangular search query time becomes $O(\log n +k)$, $O(\log n)$ for the first binary search and $O(k)$ for browsing the $k$ reported points.
\section{String Subsequence Kernels}
\label{SSK}
The philosophy of all string kernel approaches can be reduced to different ways to count common substrings or subsequences that occur in the two strings to compare. This philosophy is manifested in two steps:
\begin{itemize}
	\item Project the strings over an alphabet $\Sigma$ to a high dimension vector space $F$, where the coordinates are indexed by a subset of the input space.
	\item Compute the distance (inner product) between strings in $F$. Such distance reflects their similarity.
\end{itemize}
\paragraph{}For the String Subsequence Kernel (SSK) \cite {Lodhi:2002:TCU:944790.944799}, the main idea is to compare strings depending on common subsequences they contain. Hence, the more similar strings are ones that have the more common subsequences.
However, a new weighting method is adopted. It reflects the degree of contiguity of the subsequence in the string. In order to measure the distance of non contiguous elements of the subsequence, a gap penalty $\lambda \in ]0,1 ]$ is introduced.
Formally, the mapping function $\phi^{p}(s)$ in the feature space $F$ can be defined as follows:
\[\phi^{p}_{u}(s)=\sum_{I:u=s(I)}\lambda^{l(I)},\textrm { } u\in \Sigma^{p}. \]
The associated kernel can be written as:
\begin {eqnarray*}
K_{p}(s,t)&=&\langle \phi^{p}(s),\phi^{p}(t) \rangle\\
&=&\sum_{u\in \Sigma^{p}}\phi^{p}_{u}(s).\phi^{p}_{u}(t)\\
&=&\sum_{u\in \Sigma^{p}}\sum_{I:u=s(I)} \sum_{J:u=t(J)}\lambda^{l(I)+l(J)}.
\end {eqnarray*}
In order to clarify the idea of the SSK, we present a widespread example in the literature. Consider the strings $bar$, $bat$, $car$ and $cat$, for a subsequence length $p=2$, the mapping to the feature space is as follows:

\begin{table}[h]
\centering
\setlength{\tabcolsep}{10pt}
\begin{tabular}{c c c c c c c c c}
\hline
$\phi_{u}^{2}$ &ar&at&ba&br&bt&ca&cr&ct\\
\hline
bar &$\lambda^{2}$&0&$\lambda^{2}$&$\lambda^{3}$&0&0&0&0\\
bat &0&$\lambda^{2}$&$\lambda^{2}$&0&$\lambda^{3}$&0&0&0\\
car &$\lambda^{2}$&0&0&0&0&$\lambda^{2}$&$\lambda^{3}$&0\\
cat &0&$\lambda^{2}$&0&0&0&$\lambda^{2}$&0&$\lambda^{3}$\\
\hline
\end{tabular}
\end{table}

The unnormalized kernel between $bar$ and $ bat$ is $K_{2}(bar,bat)=\lambda^{4}$, while the normalized version is obtained by :
\[\widehat{K_{2}}(bar,bat)={K_{2}(bar,bat)}/{\sqrt{K_{2}(bar,bar).K_{2}(bat,bat)}}=\lambda^{4}/(2\lambda^{4}+\lambda^{6})=1/(2+\lambda^{2}).\]
\\
A direct implementation of this kernel leads to $O(|\Sigma ^{p}|)$ time and space complexity. Since this is the dimension of the feature space. To assist the computation of the SSK a Suffix Kernel is defined through the embedding given by:
\[\phi^{p,S}_{u}(s)= \sum_{I\in I_{p}^{|s|}:u=s(I)}\lambda^{l(I)}, u\in \Sigma^{p},\]
where $I_{p}^{k}$ denotes the set of $p$-tuples of indices $I$ with $i_{p}=k$. In other words, we consider only the subsequences of length $p$ that the last symbol is identical to the last one of the string $s$.
The associated kernel can be defined as follows:

\begin {eqnarray*}
K_{p}^{S}(s,t)&=&\langle \phi^{p,S}(s),\phi^{p,S}(t) \rangle\\
&=&\sum_{u\in \Sigma^{p}}\phi^{p,S}_{u}(s).\phi^{p,S}_{u}(t).\\
\end {eqnarray*}
To illustrate this kernel counting trick, we take back the precedent example where the mapping is as follows:

\begin{table}[h]
\centering
\setlength{\tabcolsep}{10pt}
\begin{tabular}{ccccccccc}
\hline
$\phi_{u}^{2,S}$ &ar&at&ba&br&bt&ca&cr&ct\\
\hline
bar &$\lambda^{2}$&0&0&$\lambda^{3}$&0&0&0&0\\
bat &0&$\lambda^{2}$&0&0&$\lambda^{3}$&0&0&0\\
car &$\lambda^{2}$&0&0&0&0&0&$\lambda^{3}$&0\\
cat &0&$\lambda^{2}$&0&0&0&0&0&$\lambda^{3}$\\
\hline
\hline
\end{tabular}
\end{table}

for example 
$K_{2}^{S}(bar ,bat)=0$ and 
$K_{2}^{S}(bat ,cat)=\lambda^{4}.$\\
The SSK can be expressed in terms of its suffix version as:
\begin{eqnarray}
\label{kp}
K_{p}(s,t)= \sum_{i=1}^{|s|}{\sum_{j=1}^{|t|}{K_{p}^{S}(s(1:i),t(1:j))}}.
\end{eqnarray}
with $K_{1}^{S}(s,t)= [s_{|s|}=t_{|t|}]\ \lambda^{2}.$

\subsection {Naive Implementation}
The computation of the similarity of two strings ($sa$ and $tb$) is conditioned by their final symbols. In the case where $a=b$, we have to sum kernels of all prefixes of $s$ and $t$. Hence, a recursion has to be devised:
\begin{eqnarray}
\label{KPS}
K_{p}^{S}(sa,tb)&=&[a=b] \sum_{i=1}^{|s|}\sum_{j=1}^{|t|}\lambda^{2+|s|-i+|t|-j}K_{p-1}^{S}(s(1:i),t(1:j)).
\end{eqnarray}
This computation leads to a complexity of $O(p(|s|^{2}|t|^{2}))$.
\subsection{Efficient Implementations}
We present three methods that compute the SSK efficiently, namely the dynamic programming~\cite{Lodhi:2002:TCU:944790.944799}, the trie-based~\cite{LeslieEtAl:03,Rousu:2005:ECG:1046920.1088717,Shawe-Taylor:2004:KMP:975545} and the sparse dynamic programming approaches~\cite{Rousu:2005:ECG:1046920.1088717}.
\\To describe such approaches, we use two strings $s = \text {gatta}$ and $t = \text {cata}$ as a running example.

\subsubsection{Dynamic Programming Approach.}
The starting point of the dynamic programming approach is the suffix recursion given by equation (\ref{KPS}). From this equation, we can consider a separate dynamic programming table $DP_{p}$ for storing the double sum:
\begin{eqnarray}
\label{DPP}
DP_{p}(k,l)&=& \sum_{i=1}^{k}\sum_{j=1}^{l}\lambda^{k-i+l-j}\,K_{p-1}^{S}(s(1:i),t(1:j)).
\end{eqnarray}
It is easy to see that:  $K_{p}^{S}(sa,tb)=[a=b]\,\lambda^{2}\,DP_{p}(|s|,|t|))$.\\
Computing ordinary $DP_{p}$ for each $(k,l)$ would be inefficient. So we can devise a recursive version of  equation (\ref{DPP}) with a simple counting device:

\begin{eqnarray*}
DP_{p}(k,l)=K_{p-1}^{S}(s(1:k),t(1:l)) +\lambda DP_{p}(k-1,l)+\\
\lambda DP_{p}(k,l-1) -\lambda^{2} DP_{p}(k-1,l-1).
\end{eqnarray*}

\noindent Consequently, using the dynamic programming approach (Algorithm~\ref{dynamic}), the complexity of the SSK becomes $O(p\,|s||t|)$.

\begin{algorithm} [h]
\DontPrintSemicolon
\KwIn {Strings $s$ and $t$, the length of the subsequence $p$, and the decay penalty $\lambda$}
\KwOut {Kernel values $K_{q}(s,t)=K(q): q=1,\ldots,p$}
$m \gets length(s)$\;
$n \gets length(t)$\;
$K(1:p) \gets 0$\;
\tcc{Computation of $K_{1}(s,t)$}
\For {i = 1:m}
{
\For {j = 1:n}
{
\If {s[i] = t[j]}
{
$KPS[i,j] \gets \lambda^{2}$\;
$K[1] \gets K[1] + KPS[i,j]$\;
}
}
}
\tcc{Computation of $K_{q}(s,t): q = 2, \ldots, p$} 
\For {q = 2:p}
{
\For {i = 1:m}
{
\For {j = 1:n}
{
$DP[i,j] \gets  KPS[i,j] +\lambda DP[i-1,j]  +\lambda DP[i,j-1] -\lambda^{2} DP[i-1,j-1]$\;
\If {s[i] = t[j]}
{
$KPS[i,j] \gets \lambda^{2}DP[i-1,j-1]$\;
$K[q] \gets K[q] + KPS[i,j]$\;
}

}
{
}
}
}
\caption{Dynamic SSK computation}

\label{dynamic}

\end{algorithm}

Table ~\ref{dyn_comp} illustrates the computation of the dynamic programming tables for the running example for $p = 1, 2$. The evaluation of the kernel is given by the sum of entries of the suffix table KPS:
\begin {eqnarray*}
K_{1}(s,t)&=&  6 \lambda ^{2}.\\
K_{2}(s,t)&=&  2\lambda ^{4} + 2\lambda ^{5} +\lambda ^{7}.\\
\end {eqnarray*} 
 
\begin{table}[h]
\caption{\label{dyn_comp} Suffix tables and dynamic programing tables to compute the SSK for $p=1, 2$. }
\centering
\setlength{\tabcolsep}{10pt}
\begin{tabular}{c c c c c c }
\hline
$KPS_{1}$ &g&a&t&t&a\\
\hline
c &&&&&\\
a &&$\lambda^{2}$&&&$\lambda^{2}$\\
t &&&$\lambda^{2}$&$\lambda^{2}$&\\
a &&$\lambda^{2}$&&&$\lambda^{2}$\\
\hline
\hline
$DP_{2}$ &g&a&t&t&a\\
\hline
c &0&0&0&0&\\
a &0&$\lambda^{2}$&$\lambda^{3}$&$\lambda^{4}$&\\
t &0&$\lambda^{3}$&$\lambda^{2}+\lambda^{4}$&$\lambda^{2}+\lambda^{3} + \lambda^{5}$&\\
a &&&&&\\
\hline
\hline
$KPS_{2}$ &g&a&t&t&a\\
\hline
c &&&&&\\
a &&&&&\\
t &&$\lambda^{4}$&&$\lambda^{5}$&\\
a &&&&&$\lambda^{4} + \lambda^{5}+\lambda^{7}$\\
\hline

\end{tabular}
\end{table}
\subsubsection{Trie-based Approach.}
This approach is based on search trees known as tries, introduced by E. Fredkin in 1960. The key idea of the trie-based approach is that leaves play the role of the feature space indexed by the set $\Sigma^{p}$ of strings of length $p$. In the literature, there are variants of trie-based string subsequence kernels. For instance the $(p,m)$-mismatch string kernel~\cite{LeslieEtAl:03} and restricted SSK~\cite{Shawe-Taylor:2004:KMP:975545}.
\begin{figure}[h]
\centering
\includegraphics[ height=50mm,width=40mm]{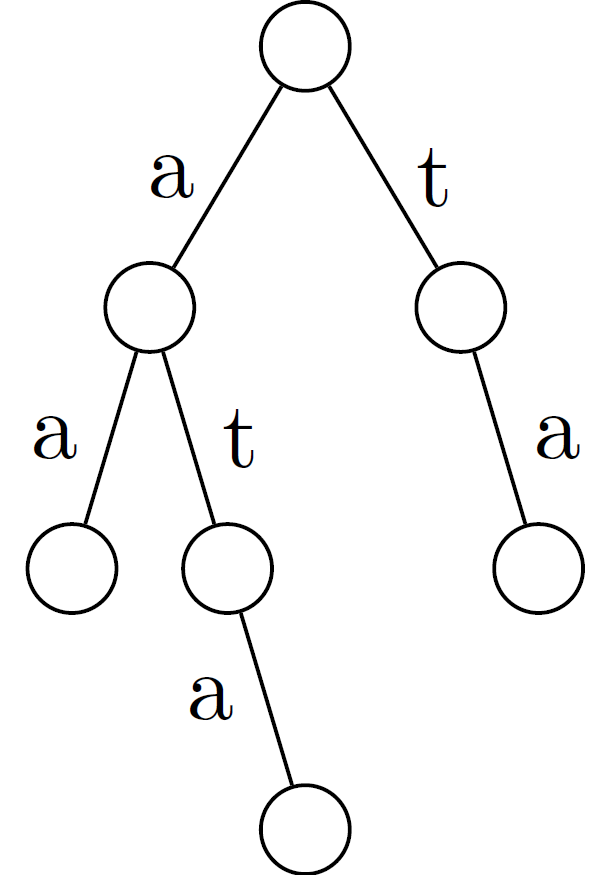}
\caption{The trie data structure for the running example $s = gatta, t = cata$ }
\label{trie}
\end{figure}
\begin{table}[h]
\caption{\label{alive} The alive indices for all subsequences of the running example for $p=1, 2, 3$ with the number of gaps from $0 \text { to } 3$.} 
\centering
\setlength{\tabcolsep}{10pt}
\begin{tabular}{ccccccc}
\hline
$g$ &$A_{s}(\text{'a'},g)$&$A_{t}(\text{'a'},g)$&$A_{s}(\text{'t'},g)$&$A_{t}(\text{'t'},g)$&$A_{s}(\text{'aa'},g)$&$A_{t}(\text{'aa'},g)$\\
\hline
0 &$\{2, 5\}$&$\{2, 4\}$&$\{3, 4\}$&$\{3\}$&&\\
1 &&&&&&$\{4\}$\\
2 &&&&&$\{5\}$&\\
3 &&&&&&\\
\hline
$g$&$A_{s}(\text {'at'},g)$&$A_{t}(\text {'at'},g)$&$A_{s}(\text {'ta'},g)$&$A_{t}(\text {'ta'},g)$&$A_{s}(\text {'ata'},g)$&$A_{t}(\text {'ata'},g)$\\
\hline
0 &$\{3\}$&$\{3\}$&$\{5\}$&$\{4\}$&&$\{4\}$\\
1 &$\{4\}$&&$\{5\}$&&$\{5, 5\}$&\\
2 &&&&&&\\
3 &&&&&&\\
\hline
\hline
\end{tabular}
\end{table}
In the present section, we try to describe a trie-based SSK presented in \cite {Rousu:2005:ECG:1046920.1088717} that slightly differ from those cited above~\cite{LeslieEtAl:03,Shawe-Taylor:2004:KMP:975545}.
Figure~\ref{trie} illustrates the trie data structure for the running example. Each node in the trie corresponds to a co-occurrence between strings. The algorithm maintains for all matches $u=s(I) = u_{1}\cdots  u_{q}$, $I=i_{1}\cdots i_{q} $ a list of alive matches $A_{s}(u,g)$ as presented in Table~\ref{alive} that records the last index $i_{q} $ where $g = l(I) - |I|$ is the number of gaps in the match. Notice that in the same list we are able to record many occurrences with different gaps. Alive lists for longer matches $uc, c \in \Sigma$, can be constructed incrementally by extending the alive list corresponding to $u$.
Similarly, the algorithm is applied to the string $t$. The process will continue until achieving the depth $p$. For efficiency reasons, we need to restrict the number of gaps to a given integer $g_{max}$, so the computation is approximate.
The kernel is evaluated as follows:
\[K_{p}(s,t)=\sum_{u\in \Sigma^{p}} \phi_{u}^{p}(s)\phi_{u}^{p}(t) =  \sum_{g_{s},g_{t}} \lambda ^{g_{s}+p}|L_{s}(u,g_{s})|\cdot\lambda ^{g_{t}+p}|L_{t}(u,g_{t})|.\]
Given that, there are $\binom{p+g_{max}}{g_{max}}$ possible combinations to assign $p$ letters and $g_{max}$ gaps in a window of length $p + g_{max}$, the worst-case time complexity of the algorithm is $O(\binom{p+g_{max}}{g_{max}}$ $(|s|+|t|))$.
\\The string subsequence kernel for the running example for $p=1$ is:
\begin {eqnarray*}
K_{1}(s,t)&=& \lambda ^{0+1}|A_{s}(\text{'a'},0)|\cdot \lambda ^{0+1}|A_{t}(\text{'a'},0)| + \lambda ^{0+1}|A_{s}(\text{'t'},0)|\cdot \lambda ^{0+1}|A_{t}(\text{'t'},0)|  \\
&=& 4\cdot \lambda ^{2} + 2\cdot \lambda ^{2} = 6\cdot \lambda ^{2}.\\
\end {eqnarray*}
\\ Similar computation is performed for $K_{2}$  and $K_{3}$:
\begin {eqnarray*}
K_{2}(s,t)&=& (1\cdot \lambda ^{2+2}) \cdot (1\cdot \lambda ^{1+2}) + (1\cdot \lambda ^{0+2} + 1\cdot \lambda ^{1+2}) \cdot (1\cdot \lambda ^{0+2}) + (1\cdot \lambda ^{0+2} + 1\cdot \lambda ^{1+2}) \cdot (1\cdot \lambda ^{0+2})\\
&=&  \lambda ^{7} + 2\cdot \lambda ^{5} + 2\cdot \lambda ^{4}.\\
\end {eqnarray*}
\\and
\begin {eqnarray*}
K_{3}(s,t)&=& (2\cdot \lambda ^{1+3}) \cdot (1\cdot \lambda ^{0+3}) = 2\cdot \lambda ^{7}.\\
\end {eqnarray*}

\subsubsection{Sparse Dynamic Programming Approach.}
It is built on the fact  that in many cases, most of the entries of the $DP$ matrix are zero and do not contribute to the result. Rousu and Shawe-Taylor~\cite{Rousu:2005:ECG:1046920.1088717} have proposed a solution using the sparse dynamic programming technique to avoid unnecessary computations. To do so, two data structures were proposed: the first one is a range sum tree, which is a B-tree, that replaces the $DP_p$ matrix. It is used to return the sum of $n$ values within an interval in $O(\log n)$ time. 
The second one is a  set of match lists instead of $K_p^S$ matrix. 
$L_{q}(i)=\{(j_{1},\overline{K_{p}^{S}}(s(1:i),t(1:j_{1})),(j_{2},\overline{K_{p}^{S}}(s(1:i),t(1:j_{2})), ...\}$
where $\overline{K_{p}^{S}}(s(1:i),t(1:j)) =\lambda ^{m-i+n-j} K_{p}^{S}(s(1:i),t(1:j))$. This dummy gap weight $\lambda ^{m-i+n-j}$ allows to address the problem of scaling the kernel values as the computation progress. Consequently the recursion~(\ref{KPS}) becomes:
\begin{eqnarray}
\label{KPSBAR}
\overline{K_{p}^{S}}(sa,tb)&=&[a=b]\lambda^{2} \sum_{i\leq |s|}\sum_{j\leq |t|}\overline{K_{p-1}^{S}}(s(1:i),t(1:j)).
\end{eqnarray}   
and the separate dynamic programming table~(\ref{DPP}) can be expressed as follows:
\begin{eqnarray}
\label{DPPBAR}
\overline{DP_{p}}(k,l)&=& \sum_{i \leq k}\sum_{j\leq l}\overline{K_{p-1}^{S}}(s(1:i),t(1:j)).
\end{eqnarray}
Thereafter, the authors devise a recursive version of~(\ref{DPPBAR}):
\begin{eqnarray}
\label{DPPBARREC}
\overline{DP_{p}}(k,l)&=& \overline{DP_{p}}(k-1,l) + \sum_{j\leq l}\overline{K_{p-1}^{S}}(s(1:i),t(1:j)).
\end{eqnarray}
This can be interpreted as reducing the evaluation of an orthogonal range query ~(\ref{DPPBAR}) to an evaluation of a simple range query multiple times as much as the number of lines of the $K_{p}^{S}$ matrix. 

To evaluate efficiently a range query, the authors use a range-sum tree to store a set $S =\{ (j,v_{j})\} \subset \{ 1,\ldots n\} \times \mathbf{R}$ of key-value pairs. A range-sum tree is a binary tree of height $h = \lceil \log {n} \rceil$ where each node in depth $d = 0,1, \ldots, h-1$ contains a key $j$ with a sum of values in a sub range $[j-2^{h-d}+1,j]$. The root is labeled with $2^{h}$, the left child of a node $j$ is $j-j/2$ and the right child if it exists is $j+j/2$. Odd keys label the leaves of the tree.

To compute the range sum of values within an interval $[1,j]$ it suffices to browse the path from the node j to the root and sum over the left subtrees as follows:
\begin{eqnarray*} 
Rangesum([1,j]) &=& v_{j} + \sum_{h \in Ancestors(j)/h < j}v_{h}.
\end{eqnarray*}
Moreover to update the value of a node $j$, we need to update all the values of parents that contain $j$ in their subtree ($h \in Ancestors(j)/h > j$). These two operations are performed in $O(\log n)$time because we traverse in the worst case the height of the tree.

For the sparse dynamic programming algorithm (Algorithm~\ref{sparse}) the range-sum tree is used incrementally when computing~(\ref{DPPBARREC}). So that when processing the match list $L_{p}(k)$ the tree will contain the values $v_{j}$ that satisfy  $\sum_{i=1}^{k}\overline{K_{p-1}^{S}}(s(1:i),t(1:j)), 1\leq j \leq l$. Hence the evaluation of~(\ref{DPPBARREC}) is performed by involving a one dimensional range query:
\begin{eqnarray*} 
Rangesum([1,j]) &=& \sum_{j=1}^{l} v_{j}\\
&=& \sum_{i=1}^{k}\sum_{j=1}^{l}\overline{K_{p-1}^{S}}(s(1:i),t(1:j))\\
&=&\overline{DP_{p}}(k,l).
\end{eqnarray*}
Concerning the cost of computation of this approach, the set of match lists is created on $O(m + n +|\Sigma| + |L_{1}|)$ time and space, while the kernel computation time  is $O(p|L_{1}| \log n)$, knowing that $|L_{1}|\geq |L_{2}|\geq \ldots \geq |L_{p}|$.

\begin{figure}[h]
\centering
\includegraphics[ height=50mm,width=40mm]{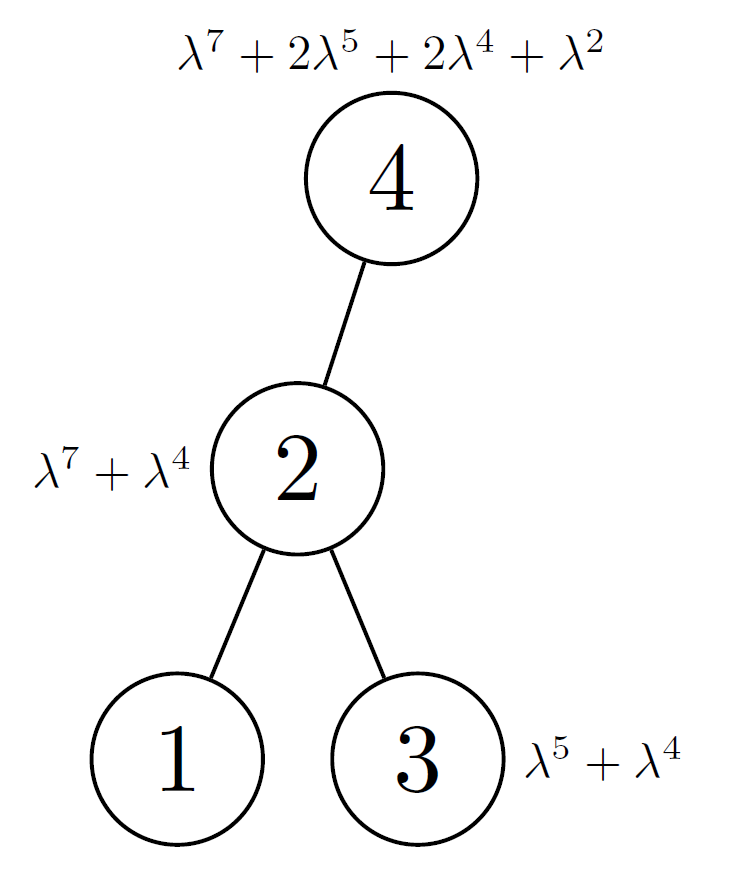}
\caption{The state of the range-sum tree when computing $K_{2}^{S}(s,t)$ }
\label{kp2s}
\end{figure}

To illustrate the mechanism of the sparse dynamic programming algorithm, Figure~\ref{kp2s} depicts the state of the range-sum tree when computing $K_{2}^{S}(s,t)$.
\\Initially the set of match lists is created as follows:
\begin{eqnarray*} 
L_{1}(1) &=& ()\\
L_{1}(2) &=& ((2,\lambda^{7}),(4,\lambda^{5}))\\ 
L_{1}(3) &=& ((3,\lambda^{5}))\\
L_{1}(4) &=& ((3,\lambda^{4}))\\  
L_{1}(5) &=& ((2,\lambda^{4}),(4,\lambda^{7})).\\ 
\end{eqnarray*}
Meanwhile maintaining the range-sum tree, the algorithm update the set of match lists as presented below:
\begin{eqnarray*} 
L_{2}(1) &=& ()\\
L_{2}(2) &=& ()\\ 
L_{2}(3) &=& ((3,\lambda^{7}))\\
L_{2}(4) &=& ((3,\lambda^{7}))\\  
L_{2}(5) &=& ((4,\lambda^{7}+\lambda^{5}+\lambda^{4})).\\ 
\end{eqnarray*}
Finally, summing the values of the updated match list after discarding the dummy weight gives the kernel value $K_{2}(s,t)$:
\begin{eqnarray*} 
K_{2}(s,t) &=& \lambda^{7} \cdot \lambda^{-9+3+3} + \lambda^{7} \cdot \lambda^{-9+4+3} +(\lambda^{7}+\lambda^{5}+\lambda^{4})\cdot \lambda^{-9+5+4} \\
&=& \lambda^{7}+2\lambda^{5}+2\lambda^{4}.\\
\end{eqnarray*}

\begin{algorithm} [h]
\DontPrintSemicolon
\KwIn {Strings $s$ and $t$,the length of the subsequence $p$, and the decay penalty $\lambda$}
\KwOut {Kernel value $K_{p}(s,t)=K$}
$m \gets length(s)$\;
$n \gets length(t)$\;
\emph{Creation of the set of match lists $L_{1}$}\;
\For {q = 2:p}
{
$Rangesum(1:n) \gets 0$ (Initialization of the range-sum tree)\; 
\For {i = 1:m}
{
\ForEach(){ $(j_{h},v_{h}) \in L_{q-1}(i)$}
{
$S \gets Rangesum[1,j_{h}-1]$\;
\If {$S > 0$}
{
appendlist$(L_{q}(i),(j_{h},S))$\;
}
}
\tcc{Update of the range-sum tree}
\ForEach(){ $(j_{h},v_{h}) \in L_{q-1}(i)$}
{
update(Rangesum, $(j_{h},v_{h})$)\;
}
}
}
\tcc{Computation of the kernel value for the final level}
$K \gets 0$\;
\For {i = 1:m}
{
\ForEach(){ $(j_{h},v_{h}) \in L_{p}(i)$}
{
$K \gets K + v_{h}\lambda ^{-m-n + i+ j_{h}}$
}
}
\caption{Sparse Dynamic SSK computation}
\label{sparse}
\end{algorithm}

\section{Geometric based Approach}
\label{LRST}
Looking forward to improving the complexity of SSK,  our approach is based on two observations. The first one concerns the computation of $K_{p}^{S}(s,t)$ that is required only when $s_{|s|}=t_{|t|}$. Hence, we have kept only a list of index pairs of these entries rather than the entire suffix table, $L(s,t) = \{(i,j): s_{i}=t_{j}\}.$

In the rest of the paper, while measuring the complexity of different computations, we will consider, $|L|$, the size of the match list $L(s,t)$ as the parameter indicating the size of the input data.

\begin{figure}[t]
\begin{center}
            \includegraphics[width=0.7\linewidth,height=0.6\textwidth,keepaspectratio]{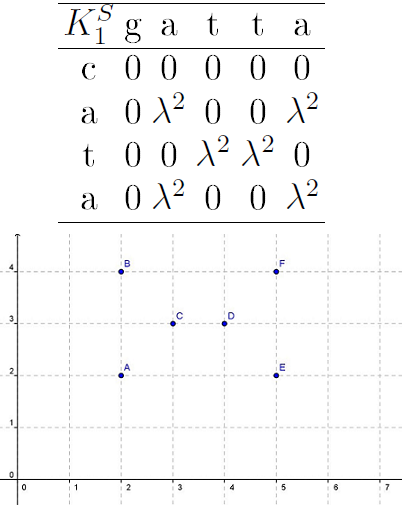}
\caption{Representation of the suffix table as a $2$-dimensional space}
\label {KPSPLANE}
\end{center}
\end{figure}

The complexity of the naive implementation of the list version is $O(p|L|^{2})$, and it seems not obvious to compute $K_{p}^{S}(s,t)$ efficiently on a list data structure.
In order to address  this problem, we have made a second observation that the suffix table can be represented as a 2-dimensional space (plane) and the entries where $s_{|i|}=t_{|j|}$ as points in this plane as depicted in Fig.~\ref{KPSPLANE}. In this case, the match list generated is \[L(s, t)=\{A, B, C, D, E, F\}= \{(2,2), (2,4),  (3,3), (4,3), (5,2),(5,4)\}.\]
With a view to improving the computation of the SSK, it is easy to perceive from Fig.~\ref{KPSPLANE} that the computation of~(\ref{KPS}) can be interpreted as orthogonal range queries. In this respect, we have used a layered range tree (LRT) in~\cite{bellaouar:efficient}. But the LRT data structure reports all points that lie in a specific range query. However, for the SSK computation we require only the sum of values of the reported points.

To achieve this goal, we extend the LRT with the aggregate operations, in particular the summation one. Hence, a novel data structure was created, for instance a Layered Range Sum Tree (LRST). 
A LRST is a LRT with two substantial extensions to reduce the range sum query time from $O(\log |L| + k)$ to $O(\log |L|)$ where $k$ is the number of reported points in the range.

The first extension consists to substitute  the associated data structures $\mathcal{T}_{assoc}$ in the LRT with new associated data structures $\mathcal{T'}_{assoc}$ where each entry $i$ contains a key-value pair $(j,ps_{j})$, $ps_{j} = \sum_{k=1}^{i} v_{k}$ is the partial sum of $\mathcal{T}_{assoc}$ in the position $i$. This extension is made to compute the range sum of $\mathcal{T}_{assoc}$ within $[i, j]$ in $O(1)$ time as follows : $Rangesum[i,j] = \mathcal{T'}_{assoc} [j] - \mathcal{T'}_{assoc} [i-1]$.

The second extension involves the fractional cascading technique. Let $\mathcal{T'}_{assoc1}$ and $\mathcal{T'}_{assoc2}$ be two sorted arrays that store partial sums of $\mathcal{T}_{assoc1}$ and $\mathcal{T}_{assoc2}$ respectively. Suppose that we want to compute the range sum within a query  $q = [y_{1}, y_{2}]$ in $\mathcal{T}_{assoc1}$ and $\mathcal{T}_{assoc2}$. 
We start with a binary search with $y_{1}$ in $\mathcal{T'}_{assoc1}$ to find the smallest key larger than or equal $y_{1}$. We make also an other binary search  with $y_{2}$ in $\mathcal{T'}_{assoc1}$ to find the largest key smaller than or equal $y_{2}$. If an entry $\mathcal{T'}_{assoc1}[i]$ stores a key $y_{i}$ then we store a pointer to the entry in $\mathcal{T'}_{assoc2}$ with the smallest key larger than or equal to $y_{i}$, say \textit {small pointer}, and a second pointer to the entry in $\mathcal{T'}_{assoc2}$ with the largest key smaller than or equal to $y_{i}$, say \textit {large pointer}. If there is no such key(s) then the pointer(s) is (are) set to \textbf {nil}.

It is easy to see that our extensions does not affect neither the space  nor the time complexities  of the LRT construction. So according to the analysis of of computational geometry algorithms, our LRST requires $O(|L|\log|L|)$ storage and can be constructed in $O(|L|\log|L|)$ time. This leads to the following lemma.
\begin{lemma}
\label{lrstconst}
Let $s$ and $t$ be two strings and $L(s,t)=\{(i,j):s_{i}=t_{j}\}$ the match list associated to the suffix version of the SSK. A Layered range sum tree (LRST) for $L(s,t)$ requires $O(|L|\log|L|)$ storage and takes $O(|L|\log|L|)$ construction time.
\end{lemma}
We can now exploit these extensions to compute efficiently the range sum inherent to $q = [y_{1}, y_{2}]$ in $\mathcal{T}_{assoc1}$ and $\mathcal{T}_{assoc2}$. Let  $\mathcal{T'}_{assoc1}[i_{1}]$ and $\mathcal{T'}_{assoc1}[i_{2}]$ be the results of the binary search with $y_{1}$ and  $y_{2}$ respectively in $\mathcal{T'}_{assoc1}$. So the $Rangesum(y_{1}, y_{2}) = \mathcal{T'}_{assoc1} [i_{2}] - \mathcal{T'}_{assoc1} [i_{1}-1]$ in $\mathcal{T}_{assoc1}$  takes $O(\log |L|)$ time. To compute the range sum in  $\mathcal{T}_{assoc2}$ we avoid the binary searches. We consider first the entry $\mathcal{T'}_{assoc2}[j_{1}]$ pointed by the \textit {small pointer} of $\mathcal{T'}_{assoc1}[i_{1}]$,  it contains the smallest key from $\mathcal{T'}_{assoc2}$ larger than or equal to  $y_{1}$. The second entry is  $\mathcal{T'}_{assoc2}[j_{2}]$ pointed by the \textit {large pointer} of $\mathcal{T'}_{assoc1}[i_{2}]$, it contains the largest key from $\mathcal{T'}_{assoc2}$ smaller than or equal to  $y_{2}$. Finally the range sum within $[y_{1}, y_{2}]$ in $\mathcal{T}_{assoc2}$ is given by $Rangesum(y_{1}, y_{2}) = \mathcal{T'}_{assoc2} [j_{2}] - \mathcal{T'}_{assoc2} [j_{1}-1]$ and it takes $O(1)$ time.

\begin{algorithm} [h]
\DontPrintSemicolon
\KwIn {Strings $s$ and $t$,the length of the subsequence $p$, and the decay penalty $\lambda$}
\KwOut {Kernel values $K_{q}(s,t)=K(q): q = 1, \ldots, p$}
$m \gets length(s)$\;
$n \gets length(t)$\;
\emph{Creation of the initial match list $L$}\;
\tcc{Computation of $K_{1}(s,t)$}
\ForEach(){ $((i,j) , v) \in L$}
{
	$K[1] \gets K[1] + v \cdot \lambda^{i+j}$\;
}
\tcc{Computation of $K_{q}(s,t): q = 2, \ldots, p$} 
\For {q = 2:p}
{
\emph{Building of the LRST corresponding to the match list $L$}\;
\ForEach(){ $((i,j) , v) \in L$}
{
\tcc{Preparing the range query for the entry $(i,j)$}
$rq \gets [(0|-\infty) :(i-1|+\infty)]\times[(0|-\infty):(j-1|+\infty)]$\;
$result \gets Rangsum[rq]$\;
\If {$result > 0$}
{
$K[q] = K[q]+ result \cdot \lambda^{i+j}$\;
appendlist$(newL,((i,j),result))$\;
}
}
$L \gets newL$\;
}
\caption{Geometric SSK computation}
\label{geometric}
\end{algorithm}
For our geometric approach (Algorithm~\ref{geometric}) we will use the LRST to evaluate the SSK. We start by the creation of the match list $L(s,t)=\{((i,j),\widetilde{K_{p}^{S}}(s(1:i),t(1:j))):s_{i}=t_{j}\}$ where $\widetilde{K_{p}^{S}}(s(1:i),t(1:j)) = \lambda^{2-i-j}K_{p}^{S}(s(1:i),t(1:j))$. This trick is inspired from \cite{Rousu:2005:ECG:1046920.1088717} to make the range sum results correct. Thus the recursion~(\ref{KPS}) becomes as follows:
\begin{eqnarray}
\label{KPStilde}
\widetilde{K_{p}^{S}}(sa,tb)&=&[a=b]\sum_{i\leq |s|}\sum_{j\leq |t|}\lambda^{i+j}\widetilde{K_{p-1}^{S}}(s(1:i),t(1:j)).
\end{eqnarray}

In order to construct efficiently the match list we have to create for each character $c \in \Sigma$ a list $I(c)$ of occurrence positions $(c=s_{i})$ in the string $s$. Thereafter, for each character $t_{j}\in t$ we insert key-value pairs $((i,j),\widetilde{K_{p}^{S}}(s(1:i),t(1:j))) $ in the match list $L(s,t)$ corresponding to $I(t_{j})$. This process takes $O(|s| + |t| + |\Sigma| + |L|)$ space and $O(|s| + |\Sigma| + |L|)$ time. For example, the match list for our running example is $L(s,t)=\{((2, 2), \lambda^{7}), ((2, 4), \lambda^{5}), ((3, 3), \lambda^{5}), ((4, 3), \lambda^{4}), ((5, 2), \lambda^{4}), ((5, 4), \lambda^{2})$.

Once the initial match list created, we start computing the SSK for the subsequence length $p = 1$. This computation doesn't require the LRST ; it suffices to traverse the match list and sum over its values. For length subsequence $q>1$ we will first create the LRST corresponding to the match list, afterward for each item $((k,l),\widetilde{K_{p}^{S}}(s(1:k),t(1:l)))$ we invoke the LRST with the query $rq = [0, k-1] \times [0, l-1]$. This latter return the range sum within $rq$:  $Rangesum(rq) = \sum_{i<k}\sum_{j<l} (\widetilde{K_{p}^{S}}(s(1:i),t(1:j)))$. If $Rangesum(rq)$ is positive then insert the key-value in a new match list for the level $q+1$ and  summing the $Rangesum(rq)$ to compute the SSK at the level $q$. At each iteration, we have to create a new LRST corresponding to the new match list until achieving the request subsequence length $p$.

We recall that in our case, we use  composite numbers instead of real numbers, see section (\ref{lrt}). In such  situation, we have to transform the range query $[x_{1}:x_{2}]\times[y_{1}:y_{2}]$ related to a set of points in the plane to the range query $[(x_{1}|-\infty) :(x_{2}|+\infty)]\times[(y_{1}|-\infty):(y_{2}|+\infty)]$ related to the composite space.

Using our geometric approach, the range sum query time becomes $O(\log|L|)$. For the computation of $K_{p}^{S}(s,t)$ we have to consider $|L|$ entries of the match list. The process iterates $p$ times, therefore, we get a time complexity of $O(p|L|\log{|L|})$ for evaluating the SSK. This result combined to that of Lemma.~\ref{lrstconst} lead to the following theorem that summarizes the result of our proposed approach to compute SSK.
\begin{theorem}
Let $s$ and $t$ be two strings and $L(s,t)=\{(i,j):s_{i}=t_{j}\}$ the match list associated to the suffix version of the SSK. A layered range sum tree  requires  $O(|L|\log|L|)$ storage and it can be constructed in $O(|L|\log|L|)$ time. With these data structures, the SSK of length $p$ can be computed in $O(p|L|\log|L|)$).
\end{theorem}
\begin{figure}[h]
\includegraphics[ height=65mm,width=120mm]{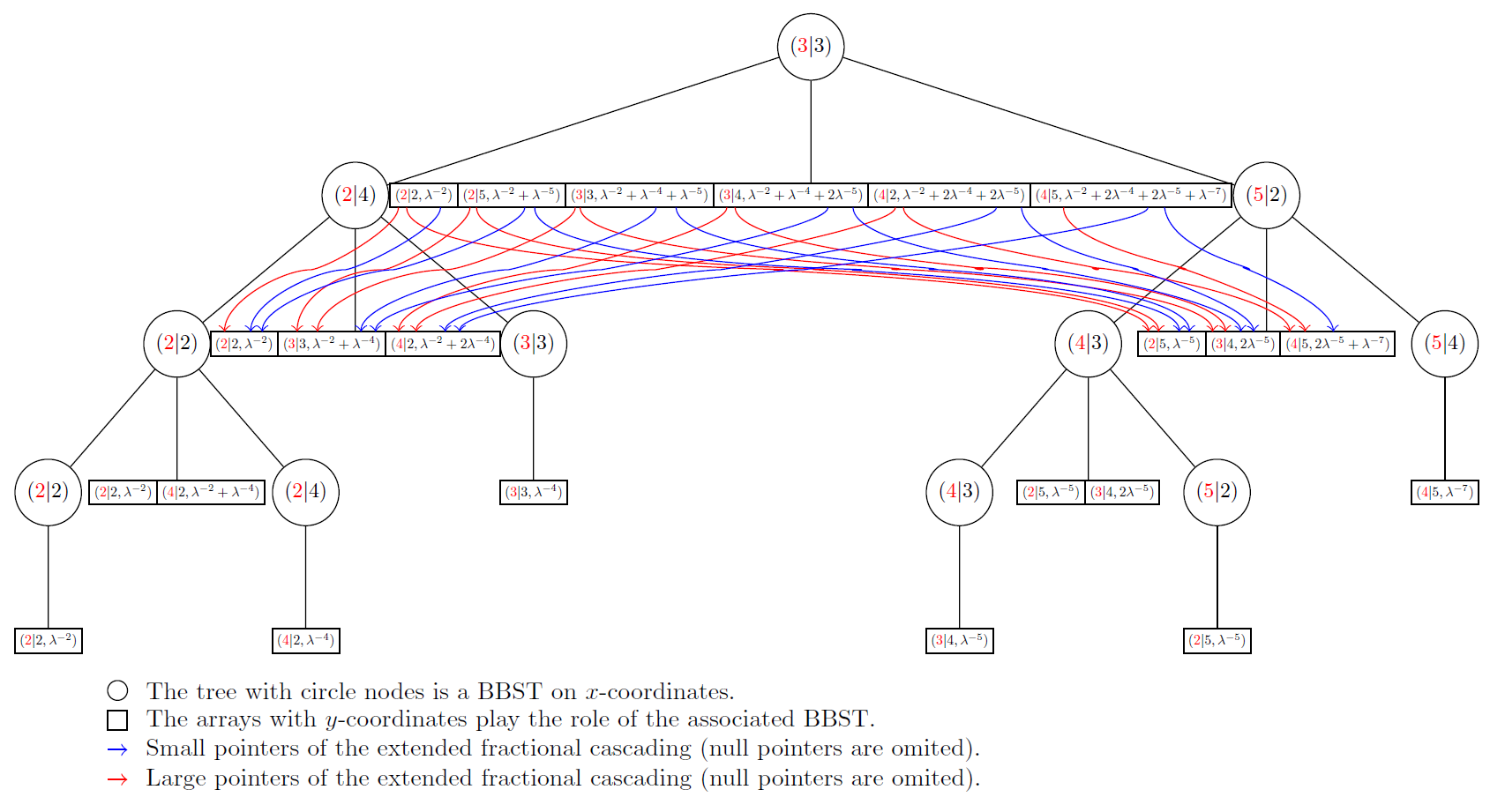}
\caption{ The state of the layered range sum tree for the running example at the step $p=2$ with an illustration of the extended fractional cascading (only between two levels).}
\label{layeredrangesumtree}
\end{figure}
To compute $K_{2}(s,t)$, for our running example, we have to invoke the range sum on the LRST at the step $p=2$ represented by Fig.\ref{layeredrangesumtree}. The SSK computation is performed by summing over all the range sums correponding th the entries of the match list as follows: $K_{2}(s,t) = Rangesum[(0|-\infty) :(1|+\infty)]\times[(0|-\infty):(1|+\infty)]+ Rangesum[(0|-\infty) :(1|+\infty)]\times[(0|-\infty):(3|+\infty)]+Rangesum[(0|-\infty) :(2|+\infty)]\times[(0|-\infty):(2|+\infty)]+Rangesum[(0|-\infty) :(3|+\infty)]\times[(0|-\infty):(2|+\infty)]+Rangesum[(0|-\infty) :(4|+\infty)]\times[(0|-\infty):(1|+\infty)] +Rangesum[(0|-\infty) :(4|+\infty)]\times[(0|-\infty):(3|+\infty)]$.

To describe how this can be processed, we deal by the range sum of the query $[(0|-\infty) :(4|+\infty)]\times[(0|-\infty):(3|+\infty)]$. At the associate data structure corresponding to the split node $(3|3)$ of Fig.\ref{layeredrangesumtree} we find the entries $(2|2)$ and $(3|4)$ whose $y-coordinates$ are the smallest one larger than or equal to $(0|-\infty)$ and the largest one smaller or equal to $(3|+\infty)$ respectively. This can be done by binary search. Next, we look for the nodes that are below the split node $(3|3)$ and that are the right child of a node on the search path to $(0|-\infty)$ where the path go left, or the left child of a node on the search path to $(4|+\infty)$ where the path go right. The collected nodes are $(3|3), (2|2), (4|3)$ and the result returned form the associated data structures is $\lambda^{-5}+\lambda^{-4}+\lambda^{-2}$. This is done on a constant time by following the small and large pointers form the associated data structure of the split node.  By the same process we obtain the following results of the invoked range sums:
\\$Rangesum[(0|-\infty) :(1|+\infty)]\times[(0|-\infty):(1|+\infty)] = 0$
\\$Rangesum[(0|-\infty) :(1|+\infty)]\times[(0|-\infty):(3|+\infty)] = 0$
\\$Rangesum[(0|-\infty) :(2|+\infty)]\times[(0|-\infty):(2|+\infty)] = \lambda^{-2}$
\\$Rangesum[(0|-\infty) :(3|+\infty)]\times[(0|-\infty):(2|+\infty)] = \lambda^{-2}$
\\$Rangesum[(0|-\infty) :(4|+\infty)]\times[(0|-\infty):(1|+\infty)] = 0$
\\After rescaling the returned values by the factor $\lambda ^{i+j}$ we obtain the value of $K_{2}(s,t)= \lambda^{-2} \cdot \lambda^{3+3}+ \lambda^{-2} \cdot \lambda^{4+3} +(\lambda^{-5} + \lambda^{-4}+\lambda^{-2}) \cdot \lambda^{5+4} = 2\lambda^{4} + 2\lambda^{5} + \lambda^{7}$. 
While invoking the range sums we will prepare the new match list for the next step. In our case the new match list contains the following matchs :  $\{((3, 3), \lambda^{-2}), ((4, 3), \lambda^{-2}), ((5, 2), \lambda^{-5}+\lambda^{-4}+\lambda^{-2})\}$.
\section{Experimentation}
\label {Experimentation}
In this section we describe the experiments that focus on the evaluation of our geometric algorithm against the dynamic and the sparse dynamic ones. Thereafter, these algorithms are referenced as Geometric, Dynamic and Sparse respectively. We have discarded the trie-based algorithm from this comparison because it is an approximate algorithm on the one hand, on the other hand in the preliminary experiments conducted in \cite{Rousu:2005:ECG:1046920.1088717}  it was significantly slower than Dynamic and Sparse.

To benefit from the empiric evaluation conducted in \cite{Rousu:2005:ECG:1046920.1088717}, we tried to keep the same conditions of their experiments. For this reason, we have conducted a series of experiments on both synthetically generated and on newswire article data on Reuter's news articles.

We ran the tests on Intel Core i7 at 2.40 GHZ processor with 16 GB RAM under Windows 8.1 64 bit. We implemented all the tested algorithms in Java. For the LRST implementation, we have extended the LRT implementation available on the page \url{https://github.com/epsilony/}.
\subsection {Experiments with synthetic data}
These experiments concern the effects of the string length and the alphabet size on the efficiency of the different approaches and to determine under which conditions our approach outperforms.

We randomly generated  string pairs with different lengths $(2,4, \ldots 8192)$ over alphabets of different sizes $(2,4, \ldots 8192)$. To simplify the string generation, we considered string symbols as integer in $[1, \text {alphabet size}]$. For convenience of data visualization, we have used the logarithmic scale on all axes. To perform accurate experiments, we have generated multiple pairs for the same string length and alphabet size and for each pair we have took multiple measures of the running time with a subsequence length $p=10$ and a decay parameter $\lambda = 0.5$.
\begin{figure}[h]
\includegraphics[ height=65mm,width=120mm]{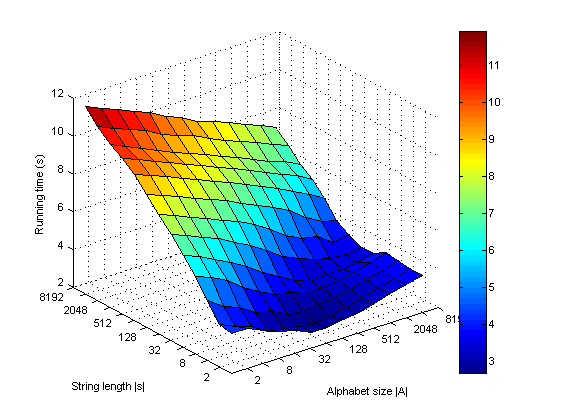}
\caption{ Running Time of the Geometric algorithm on synthetic data.}
\label{rt_geo}
\end{figure}
This being said, Fig.~\ref{rt_geo} reveals, for our geometric approach, an inverse dependency of the running time with the alphabet size. However, for an alphabet size the running time is proportional to the string length. 
\begin{figure}[h]
\centering
\includegraphics[ height=65mm,width=120mm]{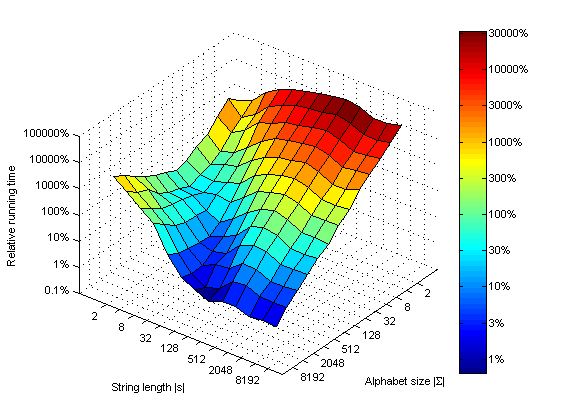}
\caption{ Relative running Time on synthetic data: Geometric/Dynamic.}
\label{rt_geo_dyn}
\end{figure}
Figure~\ref{rt_geo_dyn} shows experimental comparison of the performance of the proposed approach against Dynamic. Note that the rate of $100\%$  indicates that the two algorithms deliver the same performances. For the rates less than $100\%$ our approach outperforms, it is the case for strings based on medium and large alphabets excepting those having short length (say alphabet size great than or equal $256$, where the string length exceeds $128$ characters). For short strings and also for long strings based on small alphabets, Dynamic excels.
\begin{figure}[h]
\centering
\includegraphics[ height=65mm,width=120mm]{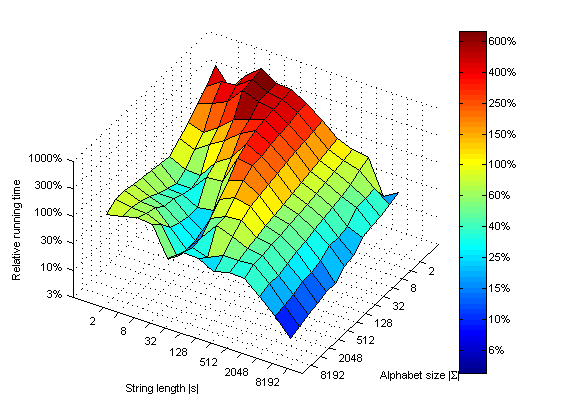}
\caption{ Relative running Time on synthetic data: Geometric/Sparse.}
\label{rt_geo_sparse}
\end{figure}
It remains to present results of the comparison experiment with Sparse which share the same motivations with our approach. Rousu and Shawe-Taylor \cite{Rousu:2005:ECG:1046920.1088717} state that with long strings based on large alphabets their approach is faster.  Figure~\ref{rt_geo_sparse} shows that in these conditions our approach dominates. Moreover, our approach is faster than the Sparse one for long strings and for large alphabets absolutely, but gets slower than Sparse for short strings based on small alphabets.
\subsection {Experiments with newswire article data}
Our second experiments use the Reuters-21578 collection to evaluate the speed of Geometric against Dynamic and Sparse on English articles. We created a dataset represented as sequences of syllables by transferring all the XML articles on to text documents. Thereafter, the text documents are preprocessed by removing stop words, punctuation marks, special symbols and finally word syllabifying. We have generated $22260$ distinct syllables. As in the first experiment, each syllable alphabet is assigned an integer. To treat the documents randomly, we have shuffled this preliminary dataset.

For visualization convenience, while creating document pairs, we have ensured that their lengths are close. Under this condition, we have collected $916$ pair documents as final dataset. 

\begin{figure}[h]
\centering
\includegraphics[ height=65mm,width=120mm]{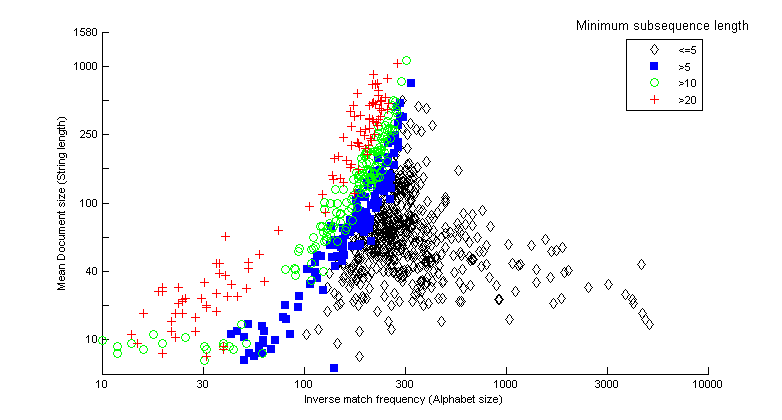}
\caption{ Clusters of document pairs where Geometric is faster than Dynamic according to the subsequence length $p$.}
\label{reuter_geo_dyn}
\end{figure}

\begin{figure}[h]
\centering
\includegraphics[ height=65mm,width=120mm]{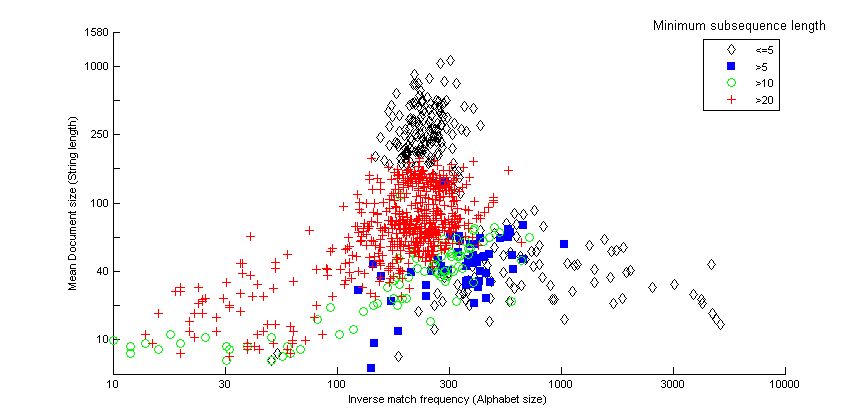}
\caption{ Clusters of document pairs where Geometric is faster than Sparse according to the subsequence length $p$.}
\label{reuter_geo_sparse}
\end{figure}

To compare the candidate algorithms, we have computed the SSK for each document pair of the data set by varing the subsequence length form $2$ to $20$.
Figure~\ref{reuter_geo_dyn} and Figure~\ref{reuter_geo_sparse} depict the clusters of documents where Geometric is faster than Dynamic and Sparse respectively.
A document pair $(s, t)$ is plotted according to the inverse match frequency (X-axis) and the document size (Y-axis). The inverse match frequency is given by: $|s||t|/|L|$, it plays the role of the alphabet size $|\Sigma|$ inherent to the documents $s$ and $t$. The document size is calculated as the arithmetic mean of the document pair sizes, it plays the role of the string length. 
Each cluster is distinguished by a special marker that corresponds to the necessary minimum subsequence length to make Geometric faster than Dynamic or Sparse. For the cluster marked by black diamonds, $p \leq 5$ is sufficient. The length $5 < p \leq 10$ is required for the cluster marked by blue filled squares. For the cluster marked by green circles $10 < p \leq 20$ is required and the last cluster marked by plus signs  $p \geq 20$ is needed.  
We can distinguish three cases in Fig.~\ref{reuter_geo_dyn}. The first one arises when the inverse match frequency is weak (small alphabet size), that is to say for dense matrix. In this case, we require important values of the subsequence length ($p > 10$ for small documents and $p > 20$ for larger ones) to make Geometric faster than Dynamic. The second case concerns good inverse match frequencies (large alphabet size) corresponding  to sparse matrix. In this case, small values of the subsequence length ($p \leq 5$) suffice to make Geometric faster than Dynamic. The third case appear for moderate  inverse match frequency (medium alphabet size), the values of $p$ that makes Geometric faster than Dynamic depend on the document size. The large document size the large $p$ is required.
The results of the comparison between Geometric and Sparse on newswire article data are depicted in Fig.~\ref{reuter_geo_sparse}. We can discuss 3 cases: The first case emerge when the document size becomes large and also for good inverse match frequency. In this case small values of the subsequence length ($p \leq 5$)  suffice to make Geometric faster than Sparse. The second case appear for small documents and bad inverse match frequencies. The necessary subsequence length must be important ($p > 10$ for very small documents and $p > 20$ for the small ones). The third case concerns modurate inverse frequencies. In this case the value of the subsequence length that makes Geometric faster than Sparse depends on the document size except large sizes which fall in the first case. 
\subsection {Discussion of the experiment results}
\label {Discussion}
In step with the results of the two experiments, it is easy to see that the algorithms behave essentially in the same way both on synthetically generated data and newswire article data. These results reveal that our approach outperforms for large alphabet size except for very small strings. Moreover, regarding to the Sparse, Geometric is competitive for long strings.

We can argue this as follows: first, the alphabet size and the string length affect substantially the kernel matrix form. Large alphabets can reduce potentially the partially matching subsequences especially on long strings, giving rise to sparse matrix form. Consequently, great number of data stored in the kernel matrix do not contribute to the result. In the other cases, for dense matrix, our approach can be worse than Dynamic by at most $Log |L|$ factor.

On the other hand, The complexities of Geometric and Sparse differ only by the factors $Log |L|$ and $Log~n$. The inverse dependency of $|L|$ and $|\Sigma|$ goes in favor of our approach. Also, the comparisons conducted on our datasets give evidence that for long strings $|L|<< n$, remembering that the size of the match list decrease while the SSK execution progress. Moreover, to answer orthogonal range queries, Sparse invoke one dimensional range query multiple times. Whereas, Geometric mark good scores by using orthogonal range queries in conjunction  with the fractional cascading and exploit our extension of the LRT data structure to get directly the sum within a range.    
\section {Conclusions and further work}
\label {Conclusions}
We have presented a novel algorithm that efficiently computes the string subsequence kernel (SSK). Our approach is refined over two phases. We started by the construction of a match list $L(s,t)$ that contains, only, the information that contributes in the result. Thereafter, in order to compute, efficiently, the sum within a range for each entry of the match list, we have extended a layered range tree to be a layered range sum tree. The Whole task takes $O(p|L|\log|L|)$ time and $O(|L|\log|L|)$ space, where $p$ is the length of the SSK and $|L|$ is the initial size of the match list.

The reached result gives evidence of an asymptotic complexity improvement compared to that of a naive implementation of the list version $O(p\,|L|^{2})$. The experiments conducted both on synthetic data and newswire article data attest that the dynamic programming approach is faster when the kernel matrix is dense. This case is achieved on long strings based on small alphabets and on short strings. Furthermore, recall that our approach and the sparse dynamic programming one are proposed in the context where the most of the entries of the kernel matrix are zero, i.e. for large-sized alphabets. In such case our approach outperforms. For long strings our approach behave better than the sparse one.

This well scaling of the proposed approach with document size and alphabet size could be useful in very tasks of machine learning on long documents as full-length research articles.

A noteworthy advantage is that our approach can be favorable if we assume that the problem is multi-dimensional. In terms of complexity, this can have influence the storage and the running time, only, by a logarithmic factor. Indeed, the layered range sum tree needs $O(|L|\log^{d-1}|L|)$ storage and can compute the sum within a rectangular range in $O(\log^{d-1}|L|)$, in a $d$-dimensional space.

At the implementation level, great programming effort is supported by well-studied and ready to use computational geometry algorithms. Hence, the emphasis is shifted to a variant of string kernel computations that can be easily adapted.

Finally, it would be very interesting if the LRST can be extended to be a dynamic data structure. This can relieve us to create a new LRST at each evolution of the subsequence length. An other interesting axis consists to combine the LRST with the dynamic programming paradigm. We believe that using rectangular intersection techniques seems to be a good track, though this seems to be a non trivial task.

\bibliographystyle{splncs03}
\bibliography{kernel}
\end{document}